\documentclass[letterpaper, 10 pt, journal, twoside]{IEEEtran}
\usepackage[pagebackref=false,breaklinks=false,colorlinks=true,bookmarks=false,linkcolor={red!50!black},urlcolor={blue},citecolor={green!60!black}]{hyperref} 

\usepackage{graphics} 
\usepackage{epsfig} 
\usepackage{mathrsfs} 
\usepackage{multirow}
\usepackage{times} 
\usepackage{amsmath} 
\usepackage{amssymb}  
\usepackage{graphicx}
\usepackage{wrapfig}
\usepackage{xcolor}
\usepackage{booktabs}

\ifCLASSINFOpdf
\else
\fi
\begin{document}
%
\title{From Prompts to Printable Models: Support-Effective 3D Generation via Offset Direct Preference Optimization}
%
%
%

\author{Chenming Wu$^{1}$, Xiaofan Li$^{1}$, and Chengkai Dai$^{1,2}$%
\thanks{Manuscript received: October, 25, 2025; Revised January, 20, 2026; Accepted February, 18, 2026.}
\thanks{This paper was recommended for publication by Editor Chao-Bo Yan upon evaluation of the Associate Editor and Reviewers' comments.
This research was partially supported by the InnoHK initiative of the Innovation and Technology Commission of the Hong Kong Special Administrative Region Government. (C. Wu and X. Li contributed equally. C. Dai is the corresponding author.)} 
\thanks{$^{1}$ Axiswise Ltd., Hong Kong SAR, China.}%
\thanks{$^{2} $ Centre for Perceptual and Interactive Intelligence, Hong Kong SAR, China.}%
\thanks{\{{\tt\footnotesize wcm94@live.com}, {\tt\footnotesize shalfunnn@gmail.com}, {\tt\footnotesize ckdai@cpii.hk}\}}
\thanks{Digital Object Identifier (DOI): see top of this page.}
}
%
%

\markboth{IEEE Robotics and Automation Letters. Preprint Version. February, 2026}
{Wu \MakeLowercase{\textit{et al.}}: Support-effective 3D Generation} 

%



\maketitle

\begin{abstract}
Current text-to-3D models prioritize visual fidelity but often neglect physical fabricability, resulting in geometries requiring excessive support structures. This paper introduces SEG (\textit{\underline{S}upport-\underline{E}ffective \underline{G}eneration}), a novel framework that integrates Direct Preference Optimization with an Offset (ODPO) into the 3D generation pipeline to directly optimize models for minimal support material usage. By incorporating support structure simulation into the training process, SEG encourages the generation of geometries that inherently require fewer supports, thus reducing material waste and production time. We demonstrate SEG's effectiveness through extensive experiments on two benchmark datasets, Thingi10k-Val and GPT-3DP-Val, showing that SEG significantly outperforms baseline models such as TRELLIS, DPO, and DRO in terms of support volume reduction and printability. Qualitative results further reveal that SEG maintains high fidelity to input prompts while minimizing the need for support structures. Our findings highlight the potential of SEG to transform 3D printing by directly optimizing models during the generative process, paving the way for more sustainable and efficient digital fabrication practices.
\end{abstract}


%
\IEEEpeerreviewmaketitle

\ifCLASSOPTIONcaptionsoff
  \newpage
\fi

\definecolor{blue}{rgb}{0,0,0}
\section{Introduction}

\textcolor{blue}{As generative models continue to mature, an increasing number of non-expert users can now create and fabricate 3D models directly from text prompts and accessible 3D printers, which becomes part of a broader democratization of personalized manufacturing. However, a significant `fabrication gap' remains: current pipelines prioritize visual fidelity over physical manufacturability. A critical bottleneck is the generation of overhanging features that necessitate excessive support structures, leading to material waste, increased print time, and surface artifacts.}

Existing text-to-3D models ignore printability constraints, while slicers handle support after geometry is fixed. These supports are crucial for maintaining the integrity of overhanging features during the printing process. Yet, they often lead to material waste and extended production times, and they can compromise the surface quality of the final product~\cite{Gao2015,Zhang2015}. Traditional geometric optimization approaches have proven effective by either deforming the 3D model in the post-processing phase~\cite{hu2015support} or decomposing the model into a sequence of subparts for multi-axis fabrication~\cite{wu2019general, dai2018support}. Unfortunately, many existing 3D generation models fail to account for these practical constraints during their training, resulting in geometries that, while visually stunning, may not be feasible for real-world fabrication.

Recognizing these challenges, we set out to create a solution that not only enhances the capabilities of 3D generation but also aligns them with the practical realities of 3D printing.
\textcolor{blue}{We name this type of pipeline as \textbf{Semantic-Fabrication Co-Optimization}, which allows the model to explore the manifold of semantically valid shapes to find one that also satisfies fabrication constraints.}
Specifically, we introduce SEG, a novel framework that employs Direct Preference Optimization with an Offset (ODPO)~\cite{amini2024direct} to refine generative models specifically for support-efficient object creation. By integrating simulation-based feedback on support structure volume into the alignment objective, SEG encourages the generation of geometries that require fewer supports while maintaining the fidelity and aesthetic appeal of the designs.
Our contributions can be summarized as follows:

\begin{itemize}
    \item We introduce SEG, the first framework that  refines 3D generative models for support-efficient design.
    \item We employ ODPO to incorporate support-effective constraints into the generation process, which effectively handles skewed reward distributions.
    \item Our experiments validate the effectiveness of SEG in a text-to-3D context, demonstrating consistent support-volume reduction and comparable geometric fidelity.
\end{itemize}

\section{Related Work}
\label{sec:related_work}

\subsection{Support-oriented Modeling and Optimization}
Support-oriented modeling and optimization methods are designed to create 3D shapes that can be fabricated with minimal or no additional support structures by embedding printability constraints directly within the modeling process. This approach addresses a fundamental challenge in 3D printing—especially in Fused Filament Fabrication (FFF)—where steep overhangs and unsupported features typically require supplementary supports to prevent collapse during printing. The use of these supports is undesirable, as it leads to increased material consumption, extended printing times, and often leaves behind rough or blemished surfaces after support removal~\cite{GAO201565}.

Prior research in support-aware modeling primarily focused on post-processing strategies that modify a completed model to reduce the need for supports or to enhance its structural stability. For example, build-orientation optimization algorithms evaluate various possible orientations of a given part, seeking to minimize the area of unsupported faces or to reduce the average angle of overhangs~\cite{dumas2014efficient}. Once an orientation is selected, explicit support-generation methods design and position support structures—ranging from tree-like branches~\cite{VAISSIER201911} to locally adaptive struts that are strategically placed according to geometric needs~\cite{Vanek14, wang2018supportnet}. These algorithms aim to uphold problematic features while balancing the trade-offs between print duration and material efficiency. Another class of techniques, partitioning-based methods, decompose complex models into smaller, more manageable sub-parts that can each be printed with fewer supports, subsequently reassembling them via geometric joints or adhesives~\cite{Luo12}.

By contrast, more recent approaches integrate support constraints directly into the modeling phase, rather than treating support generation as a secondary step. For instance, topology optimization under overhang constraints incorporates additional penalties—such as those based on feature angles or surface curvature—into classical density-based solvers, thereby discouraging the formation of unprintable regions during the design process~\cite{Gaynor16}. Other methods operate directly on the mesh geometry, applying curvature-driven smoothing to eliminate sharp overhangs, or enforcing manufacturability criteria such as minimum wall thickness and maximum unsupported span~\cite{hu2015support}.

\begin{figure*}[htbp]\centering
  \includegraphics[width=0.9\linewidth]{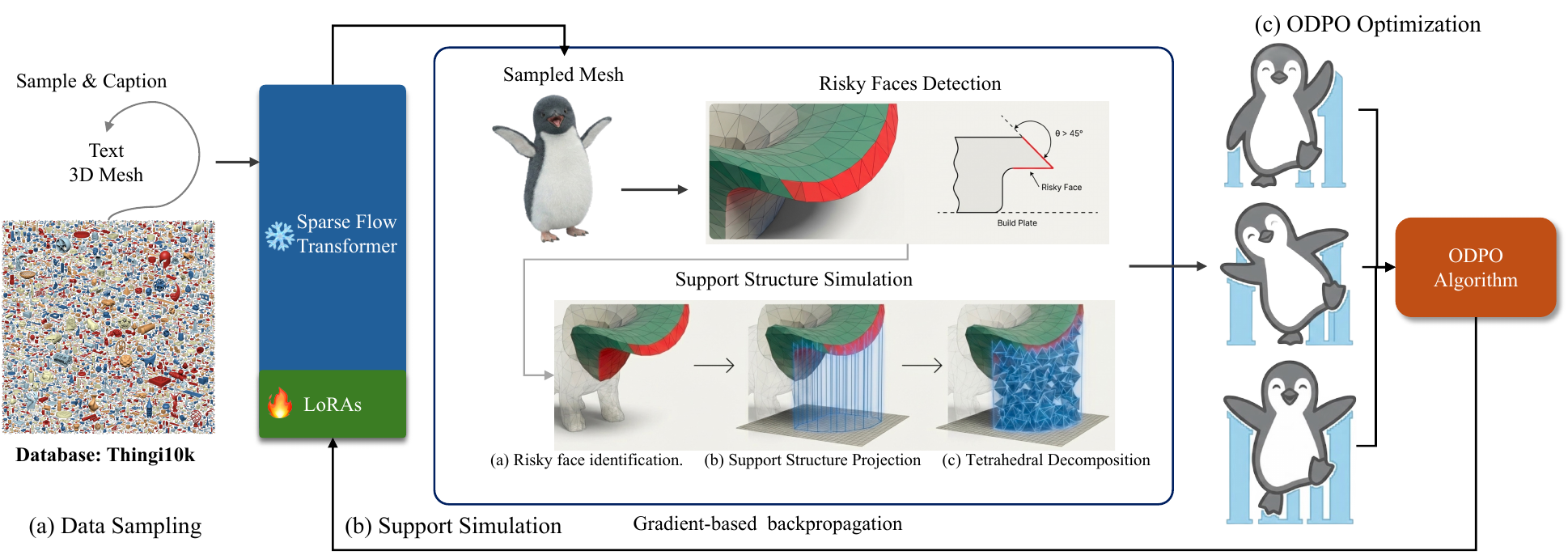}
    \caption{\textcolor{blue}{The process begins with (a) data sampling from the Thingi10k database, where multi-view images are generated and captioned using Cap3D. The Text-to-Mesh transformation utilizes a sparse flow transformer (frozen) for model generation. (b) Subsequently, support structure simulation involves estimating the support volume, identifying risky faces, and assigning NSV rewards based on overlap detection. (c) The ODPO optimization leverages the NSV rewards to optimize the LoRAs (trainable) injected in the weighted-freeze sparse flow transformer.}
    }
  \label{fig:method}
\end{figure*}
While these methods have proven effective for individual models, their reliance on case-by-case optimization or heuristic adjustment poses significant challenges for scalability.
\subsection{Text-to-3D Generation}
Text-to-3D model generation, or text-to-3D modeling, translates natural-language prompts into three-dimensional shapes by weaving semantic alignment directly into the geometry creation process. Just as support-aware modeling bakes physical manufacturing constraints into design, text-to-3D pipelines embed language-driven objectives at generation time.

One of the primary paradigms is per-prompt optimization, which sculpts a dedicated 3D representation for each input by iteratively applying Score Distillation Sampling from a frozen text-to-image diffusion model through a differentiable renderer. DreamFusion \cite{poole2022dreamfusion} pioneered this approach by refining a neural radiance field (NeRF) to match the input text. Magic3D \cite{Lin_2023_CVPR} accelerates this process with a two‐stage, coarse‐to‐fine latent diffusion pipeline and hash‐grid volumetric representation, while DreamGaussian \cite{tang2023dreamgaussian} abandons NeRFs in favor of 3D Gaussian splatting to achieve synthesis in minutes. ProlificDreamer \cite{wang2023prolificdreamer} then incorporates variational score distillation to jointly improve visual fidelity and sample diversity, and TextMesh \cite{Tsalicoglou24} adds a signed‐distance‐function backbone to produce watertight, mesh‐extractable outputs directly from text prompts. However, these methods incur heavy per-prompt computation—often on the order of minutes to hours per input—limiting interactive editing and rapid iteration.

Another key paradigm is amortized feed-forward generation, which eliminates per-prompt loops by training a single network to map text embeddings directly to 3D outputs in a single forward pass. ATT3D \cite{Lorraine23} distills multi-view diffusion guidance into a prompt-conditioned NeRF generator for one-shot synthesis. LATTE3D \cite{xie2024latte3d} scales this approach with greater capacity and diversity. GET3D \cite{gao2022get3d} trains a GAN to produce textured triangle meshes from latent codes. Shap-E \cite{Jun23} synthesizes continuous implicit fields on demand; Point-E \cite{nichol2022point} hierarchically upsamples low-resolution point clouds to full geometry. and TRELLIS \cite{xiang2024structured} introduces Structured 3D Latent representations and rectified flow transformers to generate versatile, high-quality 3D assets across multiple output formats (e.g., NeRF, Gaussian, mesh). These approaches achieve real-time throughput but focus exclusively on visual and semantic objectives.

While per-prompt methods excel in fidelity and amortized methods in speed, neither paradigm embeds physical or fabrication-aware objectives into the generation process. To address this gap, recent work integrates physical simulation or fabrication constraints directly into 3D generative pipelines to ensure stability. For example, Atlas3D~\cite{chen2024atlas3d} augments Score Distillation Sampling with a differentiable stability loss under gravity, contact, and friction to produce text-conditioned models that stand unaided in both simulation and real-world tests. Recently, Direct Simulation Optimization (DSO) utilizes non-differentiable simulator scores to label generated shapes and refines a diffusion-based 3D generator through Direct Preference Optimization for physical soundness \cite{li2025dso}. While stability or soundness can be assessed by the absolute simulated value, our focus on support efficiency aims to minimize the support structures needed for 3D models during physical fabrication. This approach encourages the generation of geometries that inherently require minimal support when produced.

\section{Proposed Method}

\textcolor{blue}{Our proposed method is built upon a pretrained diffusion-based 3D generative model $\pi_\theta(x|y)$, specifically TRELLIS~\cite{xiang2024structured}, where $y$ is a text prompt and $x$ is the latent code for a mesh $\mathcal{M}$. The Sparse Flow Transformer refers to the backbone of the first generation stage in TRELLIS, which is a diffusion transformer variant modified for flow matching on sparse voxel tokens. It determines the topology and overall shape of the generated asset. TRELLIS operates in two distinct stages, both of which utilize transformer backbones: Stage 1: Sparse Structure Generation predicts the occupancy of a 3D voxel grid, which receives a text condition and predicts a sparse set of active voxels that define the coarse geometry of the object. It is ``sparse" because it processes only the active voxel tokens rather than a dense volumetric grid, allowing for high resolution without cubic computational complexity. 
Stage 2: Structured Latent Decoding stage refines the features within the active voxels to generate detailed geometry and texture (appearance).
Our method is to align $\pi_\theta$ with the preference of support-effective printing so that the generated mesh $\mathcal{M}$ requires minimal support structure volume $V(\mathcal{M}_{sup})$ when fabricated using common 3D printing processes.}

For clarity, we begin by detailing the fundamentals of support structures in 3D printing, explaining the computational methods employed. Next, we describe how to efficiently simulate the support structure of a generated sample (mesh) using ray tracing and geometric approximation. We then discuss the preference alignment method we proposed, followed by an overview of the preference data construction pipeline we developed for training our network. An overview of our proposed SEG can be referred to Fig.~\ref{fig:method}.


\subsection{Support Structure Simulation}

In 3D printing, a model $\mathcal{M}$ is deposited layer by layer along a specified printing direction $\mathbf{d}$. Following the definitions of the maximal self-support angle $\alpha_{max}$ and self-supported regions as \textcolor{blue}{described in~\cite{hu2015support, wu2017robofdm, wu2019general, vanek2014clever,jang2020free}}, we categorize the faces of $\mathcal{M}$ based on their printability. \textcolor{blue}{For every face $f$ in the mesh, calculate the dot product of its normal $\mathbf{n}$ with the gravity vector $\mathbf{g} = (0, 0, -1)$. If $\mathbf{n} \cdot \mathbf{g} < \cos(135^\circ)$ (assuming $45^\circ$ overhang threshold), the face is risky. }

\begin{figure}[!t]\centering
  \includegraphics[width=0.7\linewidth]{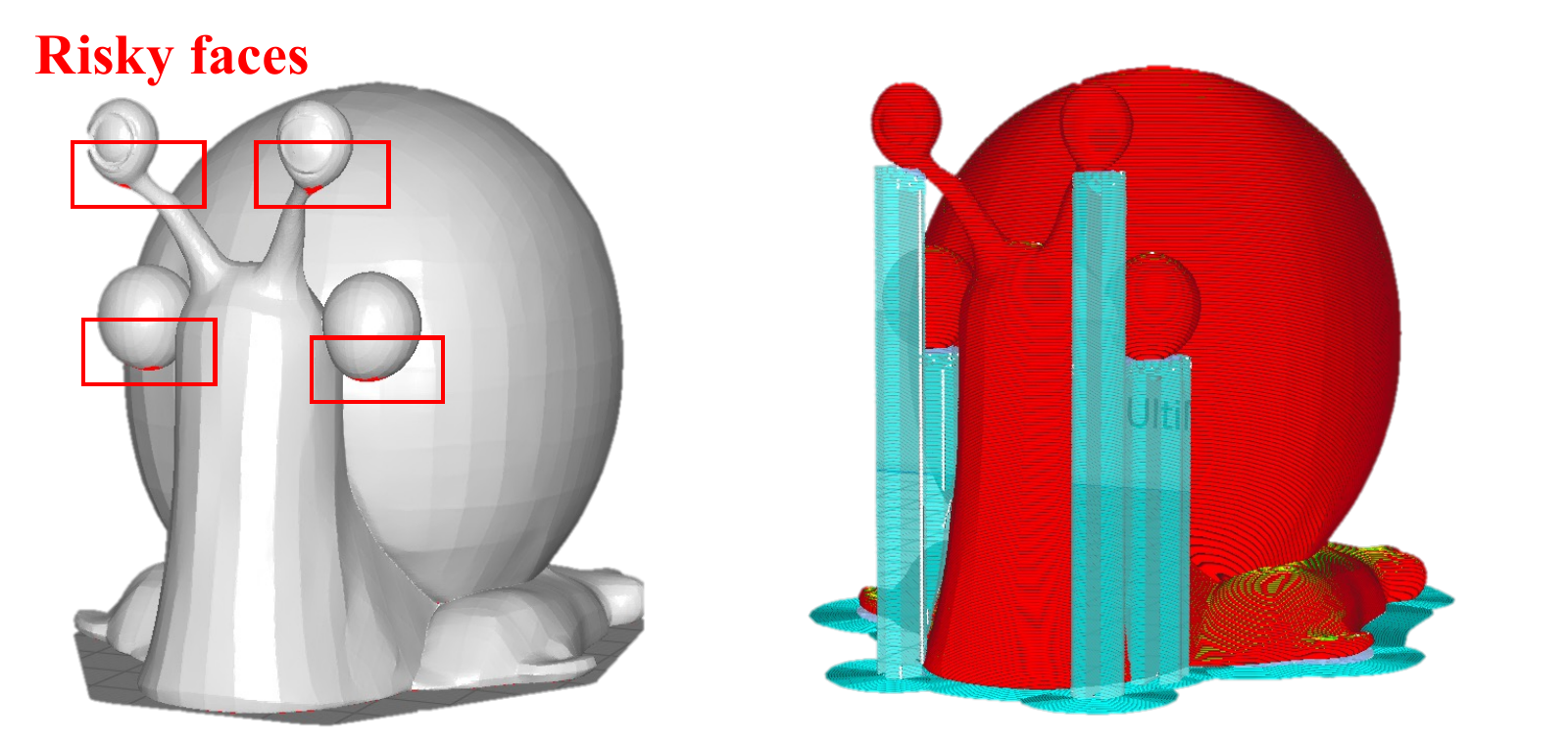}
    \caption{An example of the generated mesh featuring risky faces (left) alongside the sliced visualization, highlighting areas that require additional support in blue (right).}
  \label{fig:support}
\end{figure}

An illustrative example of risky face identification and support generation is presented in Fig.~\ref{fig:support}. The volumes of \(\mathcal{M}\) and \(\mathcal{M}_{sup}\) can be efficiently computed as \(V(\mathcal{M})\) and \(V(\mathcal{M}_{sup})\), respectively. However, since the volumes of different meshes can vary significantly, making fair comparisons of support effectiveness is challenging. To address this issue, we propose a simple yet effective metric called the normalized support volume (NSV). \textcolor{blue}{NSV serves as a scale-invariant proxy for the ``printability cost''}, which normalizes the support volume by the volume of the mesh. Specifically, NSV is defined as:

\begin{equation}
\text{NSV}(\mathcal{M}, \mathcal{M}_{sup}) = \frac{V(\mathcal{M}_{sup})}{V(\mathcal{M})}
\end{equation}

\textcolor{blue}{
From the centroid of each risky face, cast a ray in direction $\mathbf{g}$. We use a ray-tracing kernel~\cite{wald2014embree} to find the first intersection point $P_\text{int}$. If $P_\text{int}$ is on the build plate ($z=0$), the support height is $z_\text{face}$. If $P_\text{int}$ is on the mesh itself (internal support), the support height is $z_\text{face} - z_{P_\text{int}}$. This per-face ray casting is robust to non-manifold geometry (holes, self-intersections) because it does not require a watertight solid to calculate volume. 
}

\textcolor{blue}{The volume under a triangular face projected to a plane forms a triangular prism. Computing the volume of an arbitrary prism directly can be numerically unstable if the top and bottom bases are not perfectly parallel. Decomposing the prism into three tetrahedra is a standard method in Finite Element Method (FEM)~\cite{norrie2014finite} and computational geometry to calculate volumes of irregular polyhedra exactingly.$$V_\text{prism} = V_\text{tet1} + V_\text{tet2} + V_\text{tet3}$$where the volume of a tetrahedron with vertices $\mathbf{a}, \mathbf{b}, \mathbf{c}, \mathbf{d}$ is $\frac{1}{6} |(\mathbf{a}-\mathbf{d}) \cdot ((\mathbf{b}-\mathbf{d}) \times (\mathbf{c}-\mathbf{d}))|$.}

It is worth noting that our approach is agnostic to the specific support simulation method. In essence, the support simulation \( S\) proceeds by (1) identifying risky faces, (2) locating their intersection points with either the print bed or the mesh surface, and (3) accumulating the volumes of all resulting tetrahedra.

\subsection{Formulating Alignment with ODPO}
The primary objective of this work is to guide the generative process such that models requiring minimal support structures are preferentially favored. In other words, we aim to align the generation objective with the reduction of support volume: the fewer support structures a model necessitates, the higher the score or reward it should receive during evaluation. Given a set of generated 3D models' latents $\mathcal{X}=\{x_0, ..., x_n\}$ and the corresponding text prompts $\mathcal{P}=\{y_0, ..., y_N\}$, the support structure required to print it along its upright direction can be simulated as $S(x_0)$, which can be either represented by the volume of the supports or the area of the risky faces. However, no matter which representation we choose, it is an absolute metric used for geometric computing. A straightforward method following~\cite{li2025dso} is to formulate the alignment similar to Diffusion-DPO~\cite{wallace2024diffusion}. That is, we first enumerate the above simulated samples to a list of pairwise win-loss tuples $\mathcal{X}^{\text{pair}}$, any pair of them is a tuple of $(x^{\text{w}}, x^{\text{l}})$, where $w$ denotes win and $l$ denotes lose. Following~\cite{wallace2024diffusion}, the training loss is defined as follows. 

\begin{equation}
    \mathcal{L}_\text{DPO} = -\mathbb{E}_{y\sim\mathcal{P},  (x^{\text{w}}, x^{\text{l}})\sim\mathcal{X}}[\text{logsigmoid}(r_\theta(x^{\text{w}})-r_\theta(x^{\text{l}})]
    \label{eq:loss_dpo}
\end{equation}
where $r$ is the reward model.


This DPO paradigm can be trained using the samples generated by our simulator. However, upon analyzing the data distribution in the dataset, we find that training the model using the difference in NSV results in uniformly obtained samples, most of which are clustered closely together. In other words, our alignment task poses challenges due to the reward values not being constrained to the range of \(0\) to \(1\); they can either be significantly large or very small, often close to each other. This lack of diversity in the training samples does not provide sufficient information for the network to converge effectively.


\textcolor{blue}{Standard DPO struggles when the preference margin between sample pairs is small or when the reward distribution is heavily skewed. To address this, we employ ODPO~\cite{amini2024direct}, which injects a dynamic margin $\Delta o$ into the loss function. This offset is proportional to the difference in NVS between the winning and losing samples. This explicitly penalizes the model more heavily when it fails to distinguish between a highly printable mesh and a support-heavy one, effectively scaling the gradient updates based on the magnitude of the geometric improvement.}
To effectively optimize our model, we employ ODPO~\cite{amini2024direct}, which involves introducing an offset \( \Delta o \) that accounts for the differences in the reward values between preferred and dispreferred samples. Specifically, we define \( \Delta o \) based on the extent of preference indicated by the NSV values \( r(y, x^w) \) and \( r(y, x^l) \) for the preferred response \( x^w \) and dispreferred response \( x^l \), respectively. 
\begin{equation}
\Delta o = \alpha f\left(r'(y, x^w) - r'(y, x^l)\right)
\end{equation}
where \( \alpha \) is a hyperparameter that controls the extent to which the offset should be enforced, and \( f \) is a monotonically increasing function, and we propose to use \( f = \log \) in our method. Here we use the NSV metric to compute $r'$.
The loss function for ODPO can then be formulated as:
\begin{equation}
\begin{aligned}
L_{\text{ODPO}}(\theta) &= - \mathbb{E}_{y\sim\mathcal{P},  (x^{\text{w}}, x^{\text{l}})\sim\mathcal{X}} \\ & \left[ \log \sigma\left(r_\theta(y, x^w)  - r_\theta(y, x^l) - \Delta o\right)\right]
\end{aligned}
\label{eq:loss_odpo}
\end{equation}
where \( r_\theta(x, y) \) represents the estimated reward for the response given the prompt.
By applying this loss function, we ensure that the model increases the likelihood of the preferred response \( x^w \) over the dispreferred response \( x^l \) by an offset that reflects their relative quality.

This approach enhances our model's ability to align with human preferences more accurately, particularly in scenarios where the reward values vary widely. By prioritizing larger differences in rewards while still learning from smaller distinctions, ODPO significantly improves the fine-tuning process of our DPO model.

\begin{figure*}[!t]\centering
  \includegraphics[width=0.95\linewidth]{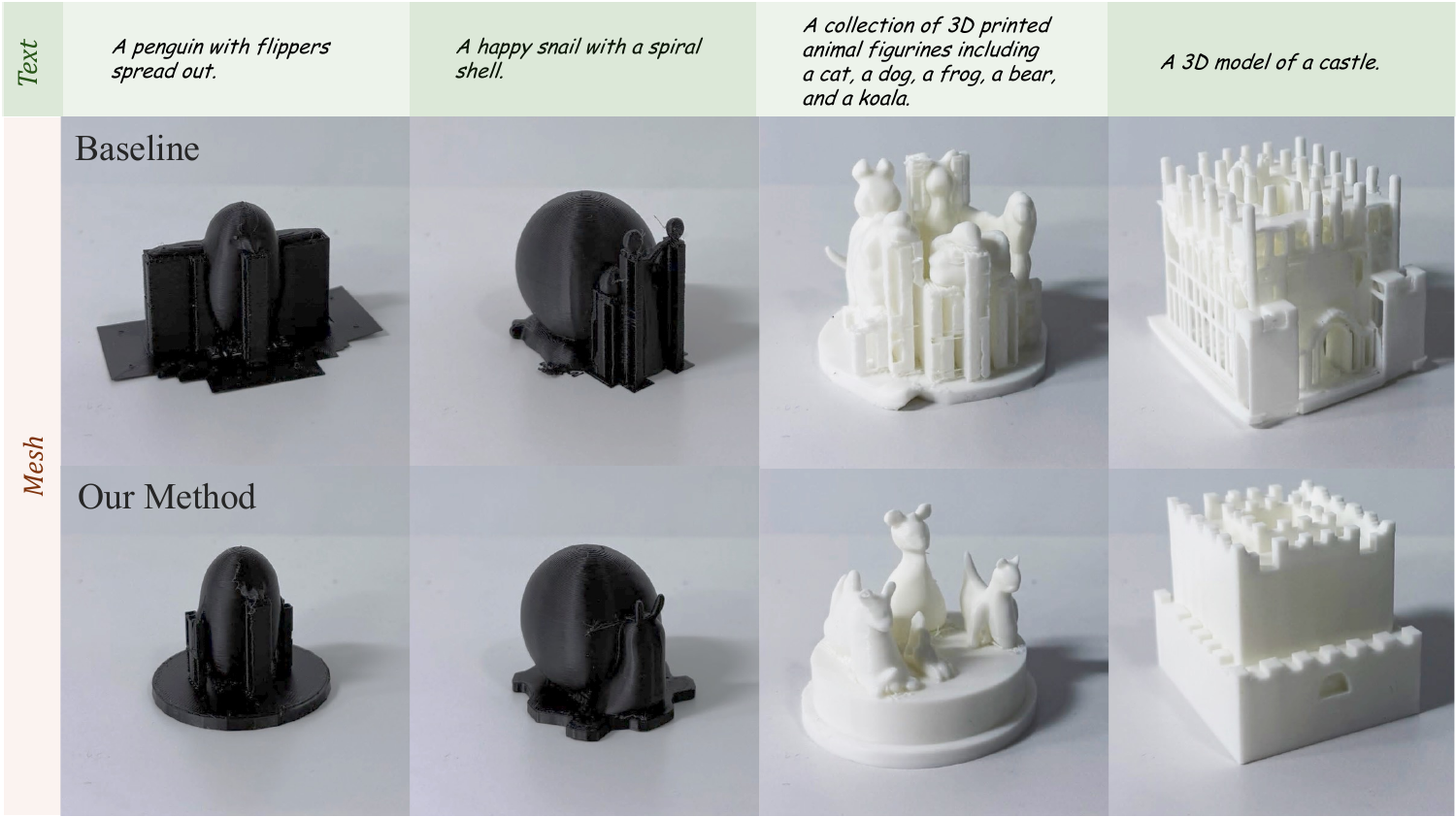}
  \caption{Physical fabrication comparison using Fused Deposition Modeling (FDM). Top: Baseline (TRELLIS) outputs require dense support structures (blue regions in simulation) to support overhangs. Bottom: Our SEG framework generates self-supporting geometries that retain the input prompt's semantics while minimizing material waste.}
  \label{fig:teaser}
\end{figure*}

\subsection{Preference Construction}
Our work specifically addresses the application of 3D printing (3DP), utilizing the Thingi10k dataset \cite{zhou2016thingi10k}, which is widely recognized in the 3DP research community. Notably, some raw data in the \cite{zhou2016thingi10k} dataset contain non-manifold faces, leading to inconsistencies in data curation. To preprocess the dataset, we employ the method described in \cite{hu2020fast}, which involves tetrahedralization to extract manifold surfaces.

For data curation in our training process, we begin by rendering multi-view images of models from the Thingi10k dataset, generating eight distinct views for each model to aid in text captioning. We employ Cap3D \cite{cap3d} to create high-quality and detailed captions with minimal hallucination, resulting in a total of \(9,840\) prompt-model pairs. Subsequently, we utilize the methodology outlined in \cite{xiang2024structured} to sample ten different latent representations for each prompt. Finally, we apply the aforementioned support structure simulation approach to determine the requisite support volume and NSV for training purposes.

\section{Results}

\subsection{Experimental Setup}
\paragraph{Implementation Details.}
\textcolor{blue}{
In our SEG framework, the objective is to alter the geometry to reduce supports. Therefore, the LoRA adapters must be injected into the Sparse Flow Transformer. We train our model using $8\times$ NVIDIA A100 GPUs, each of which has 40GB VRAM, leveraging DeepSpeed~\cite{rasley2020deepspeed} with ZeRO-1 offload for enhanced performance. The learning rate is configured to \(5 \times 10^{-6}\), and we utilize a batch size of 96 per iteration through gradient accumulation, completing a total of 50,000 training steps with the Adam optimizer~\cite{kingma2014adam}. Our foundation is the state-of-the-art 3D diffusion-based generative model TRELLIS (text-to-3D, XL size) \cite{xiang2024structured}, which serves as a robust starting point.
To efficiently validate our proposed method, LoRA adapters~\cite{hu2022lora} are attached to the Self-Attention and Cross-Attention layers of the transformer blocks. Specifically, we target the projection matrices for Query ($Q$), Key ($K$), Value ($V$), and Output ($O$).
A rank of $r=64$ is typically sufficient for geometric adaptation in diffusion transformers, therefore, we specificy the rank as 64 and alpha as 128. For a pre-trained weight matrix $W_0 \in \mathbb{R}^{d \times k}$, the adapted weight is $W = W_0 + \Delta W = W_0 + BA$, where $B \in \mathbb{R}^{d \times r}$ and $A \in \mathbb{R}^{r \times k}$ are the trainable low-rank matrices. For ODPO, the $\alpha$ is set to $1$. This approach enables effective fine-tuning using $\mathcal{L}_\text{ODPO}$ (Eq.~\ref{eq:loss_odpo}) while preserving the model's original capabilities. 
}
\paragraph{Benchmark Datasets} 
To validate the effectiveness of our proposed method in terms of support-efficient 3D printing (3DP), we created two benchmark datasets. The first benchmark dataset, named \texttt{Thingi10k-Val}, is derived from the aforementioned curated dataset, where we split the entire Thingi10k~\cite{zhou2016thingi10k} dataset into training and validation sets. The validation set consists of 100 prompts randomly selected from this split.
We also utilize GPT-4.1 to generate 60 prompts specifically designed for 3D generation in daily life, animation, and animal categories. These prompts were selected with a focus on scenarios likely to require support structures, ensuring that we account for cases that are too straightforward and do not necessitate supports, such as a simple cube. This dataset is referred to as \texttt{GPT-3DP-Val} in the following context.

\paragraph{Evaluation Metrics}
We utilize the NSV metric to evaluate the support structures generated by our proposed method and the baseline models, as it directly relates to the goal of support-efficient 3D generation. To extend NSV metric from a single model to the whole dataset, we define the volume-weighted NSV as the default computation method, and denote the arithmetic mean of NSV values as NSV$^*$. Additionally, different models may yield varying results for the same prompt, allowing us to assess the rate considering the cases of both win and tie (under a threshold of $10^{-3}$) based on NSV across the entire benchmark dataset, a measure we refer to as support-efficient consistency (SEC). \textcolor{blue}{To evaluate the semantic fidelity, we computed CLIP scores~\cite{radford2021learning} for the generated meshes.}

\paragraph{Baselines}
Our model is built upon a pretrained diffusion-based 3D generation framework, which provides a solid foundation for our approach. We specifically select the pretrained model, referred to as TRELLIS~\cite{tang2023dreamgaussian}.
In addition to this strong baseline, we train two more models using vanilla Direct Preference Optimization (DPO) and the Direct Reward Optimization (DRO)~\cite{li2025dso}. These comparisons are primarily aimed at assessing physical stability, enabling us to evaluate the effectiveness of our approach in enhancing the robustness and reliability of the generated 3D models.

\subsection{Quantitative Results}

\begin{table}[]
\centering
\caption{{Quantitative evaluation across different baselines on \texttt{Thingi10k-Val}.}} 
\begin{tabular}{@{}lllll@{}}
\toprule
\multirow{2}{*}{Method} & \multicolumn{4}{c}{\texttt{Thingi10k-Val}} \\ \cmidrule(l){2-5} 
                        & NSV$\downarrow$    & NSV$^*\downarrow$    & SEC$\uparrow$  & CLIP $\uparrow$ \\ \midrule
TRELLIS                &       0.343             &          1.255            &            N/A  & \textbf{22.13}   \\
DPO                     &   0.310                   &          1.200            &         0.740      & 21.99  \\
DRO                     &      5.999              &     18.467                 &     0.070       & 20.50     \\
\midrule
\textbf{Ours}           &    \textbf{0.176}                &           \textbf{0.587}           &       \textbf{0.870}      & 22.01     \\ \bottomrule
\end{tabular}
\label{tab:thingi10k}
\end{table}

\begin{table}[]
\centering
\caption{{Quantitative evaluation across different baselines on \texttt{GPT-3DP-Val}.}} 
\begin{tabular}{@{}lllll@{}}
\toprule
\multirow{2}{*}{Method} & \multicolumn{4}{c}{\texttt{GPT-3DP-Val}} \\ \cmidrule(l){2-5} 
                        & NSV $\downarrow$    & NSV $^*\downarrow$    & SEC $\uparrow$   & CLIP $\uparrow$ \\ \midrule
TRELLIS                &        0.504            &      1.265                &      N/A   & \textbf{26.37}        \\
DPO                     &      0.432              &         1.021             &     0.733      & 25.14      \\
DRO                     &     6.128               &      19.642                &    0.067        & 22.12     \\ \midrule
\textbf{Ours}           &       \textbf{0.222}             &     \textbf{0.691}                &        \textbf{0.867}      & 25.89   \\ \bottomrule
\end{tabular}
\label{tab:gpt}
\end{table}

We conducted experiments on two validation datasets. The results of these evaluations are summarized in Tab.~\ref{tab:thingi10k} and Tab.~\ref{tab:gpt}. Our findings reveal that our method consistently outperforms all other evaluated techniques across both datasets. 
\textcolor{blue}{The DRO baseline exhibited a performance collapse. Unlike preference-based methods, which are anchored to a reference policy, DRO directly maximizes the reward. In the context of text-to-3D, this unconstrained optimization caused the model to drift into degenerate geometries that maximized volume or created unprintable artifacts, confirming that reference-guided alignment is essential for preserving semantic fidelity while optimizing for physics.}

\begin{figure}[!t]\centering
  \includegraphics[width=1.0\linewidth]{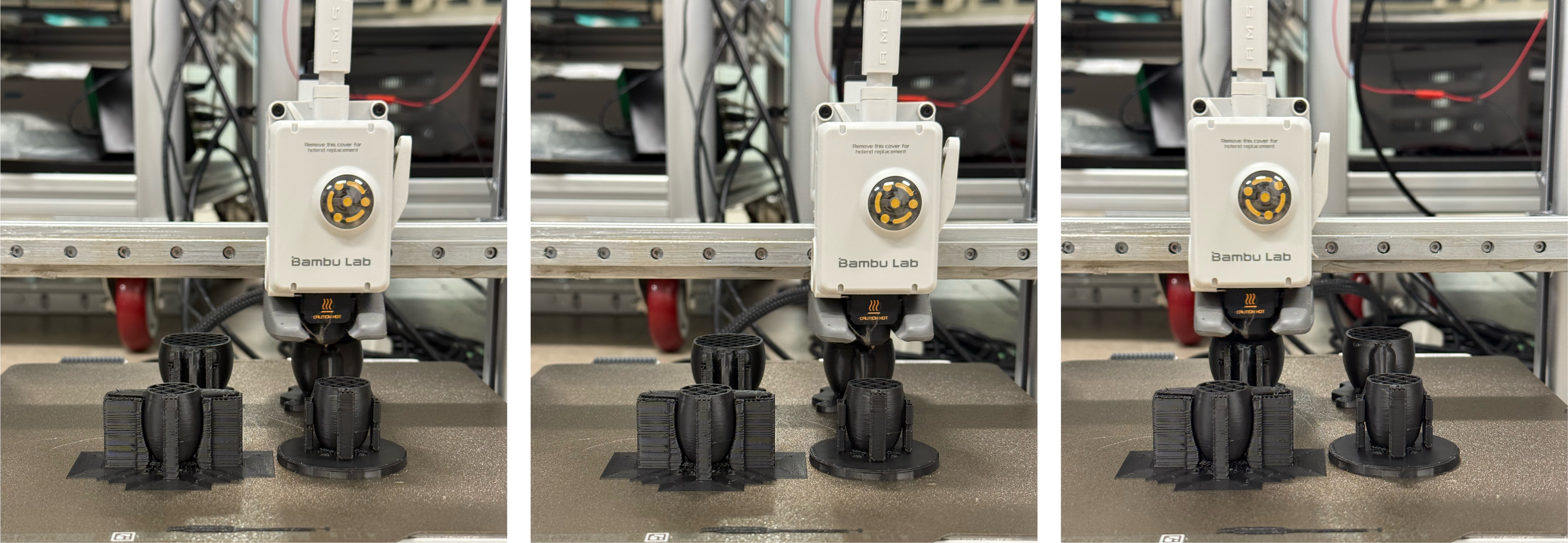}
    \caption{3D printing process using the Bambu Lab A1 printer for models generated by both the baseline diffusion model and our optimized approach.}
  \label{fig:printing}
\end{figure}

\begin{figure}[!t]\centering
  \includegraphics[width=\linewidth]{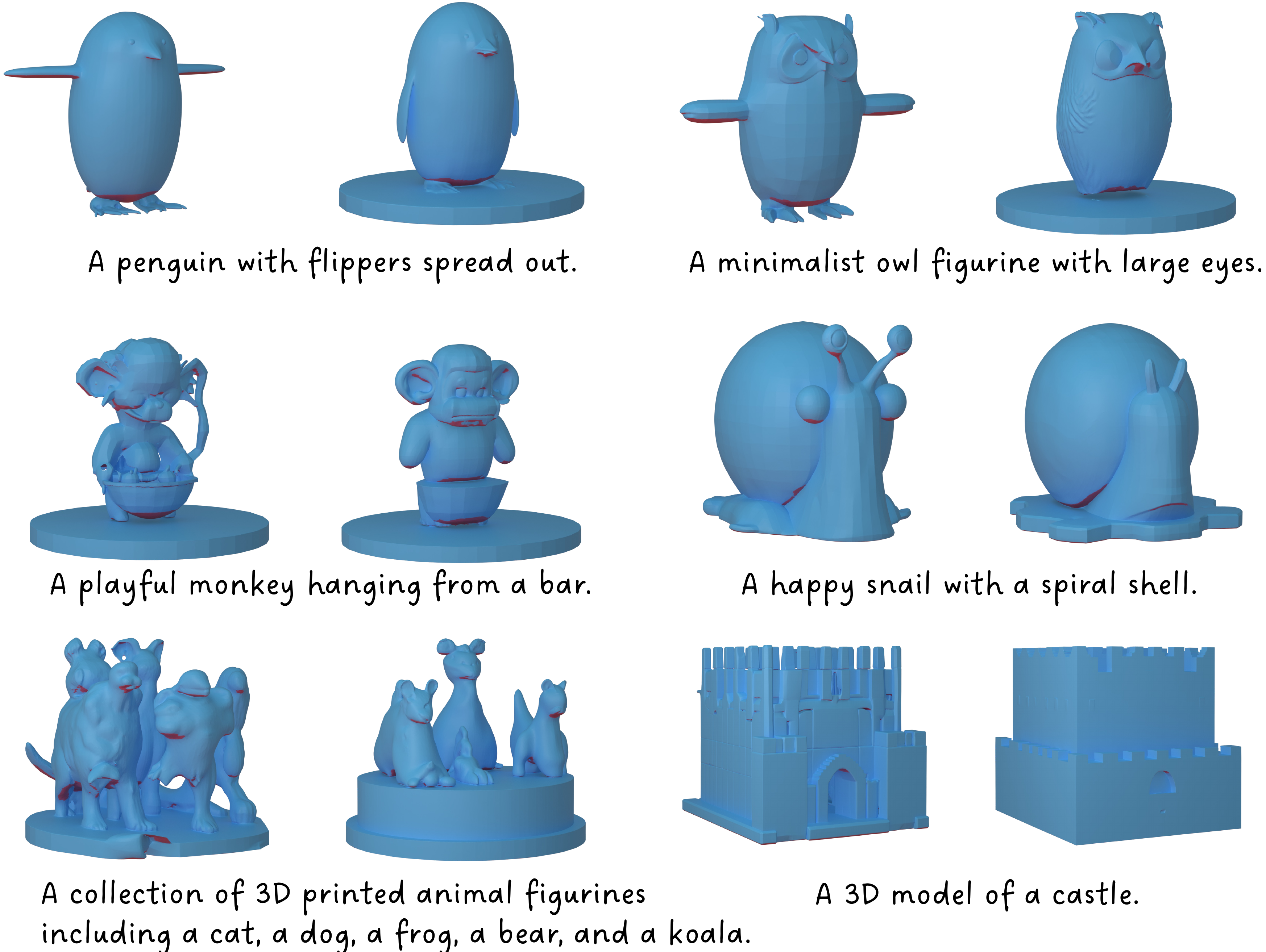}
    \caption{\textcolor{blue}{More visualization results generated by TRELLIS~\cite{xiang2024structured} (left) and our model (right). The models are colored in blue, while the risky areas are highlighted in red.}}
  \label{fig:more_results}
\end{figure}
We present visualizations of the results generated by our method in comparison to the baseline model in Fig.~\ref{fig:more_results}. To enhance clarity, we have highlighted the areas deemed risky in red, indicating regions that necessitate the addition of support structures to prevent the collapse of overhanging sections during the printing process.
Furthermore, we have fabricated several pairs of results using the Bambu Labs A1 3D printer, as depicted in Fig.~\ref{fig:printing}. This practical demonstration, featured in Fig.~\ref{fig:teaser}, underscores the potential applications of our method in real-world physical printing scenarios. 

\textcolor{blue}{A key advantage of SEG is that it effectively learns by the NSV metric, which can measure relative differences, rather than requiring absolute precision. We validated this premise by performing a correlation analysis between our NSV metric and the support material volume calculated by the commercial slicer Cura. By comparing the weight ratios (with and without supports) of 50 models sampled from Thingi10k using Cura's default `Normal' settings, we observed a positive Pearson correlation coefficient ($r > 0.85$), which confirms the effectiveness of our approximation.}


\begin{table}[h]
    \centering
    \caption{{Statistics of the ablation study.}} 
    \label{tab:ablation_track}
    \begin{tabular}{@{}llll@{}}
\toprule
            & \multicolumn{1}{c}{NSV$\downarrow$} & NSV$^*\downarrow$ & SEC$\uparrow$ \\ \midrule
$\alpha=2$             &      0.516                               &    3.468               &         0.590      \\
w/o offset          &            0.310                         &          1.200         &      0.740         \\  
early stop          &               0.242                      &           0.864        &       0.790        \\ \midrule
{full} &                   \textbf{0.176}                  &       \textbf{0.587}            &        \textbf{0.870}       \\ \bottomrule
\end{tabular}
\label{tab:ablation}
\end{table}

\subsection{Ablation Study}
We perform an ablation study to evaluate the impact of various design choices. The quantitative results of our experiments are presented in Tab.~\ref{tab:ablation}, including a larger value of $\alpha=2$, the absence of an offset, and an early stopping criterion at half the total iterations (\textit{i.e.,} 25K). 
The results clearly demonstrate that our full model achieves the best performance across all three metrics: NSV, NSV$^*$, and SEC. This highlights the significance of our design choices.

\subsection{Discussions}

Our approach is a preliminary effort to fine-tune a diffusion-based 3D generative model for specific applications, though some limitations remain. The reliance on the manifold assumption for simulating support structures may introduce biases, especially when dealing with double-sided meshes or floating facets. These issues are not inherent to our method and could be addressed with future geometry-assured 3D generation models.
\textcolor{blue}{The problem of modifying existing designs to be largely orthogonal to our generative workflow, where both can serve overlapping user groups. Our method could be extended to other generative settings in the future. The training cost of our method follows the ``train once, run everywhere'' nature of foundation models. Each inference takes only ~50 seconds on our single A100 GPU. Since our SEG is fine-tuned from a pretrained 3D generative model, some failure cases arise from limitations inherited from the underlying model. In such cases, it is difficult to clearly attribute the failures specifically to our attempt to minimize support structures.}
While SEG primarily targets FDM printers with a standard self-support angle, it is flexible and can be extended to other 3D printing technologies like SLA and SLS. Future work will explore optimizing upright orientation and self-support angles to enhance support efficiency across different printing methods.



\section{Conclusion}
This paper introduces the SEG framework, which uses ODPO to enhance 3D generative models for support-efficient printing. By addressing the critical issue of support structures, often overlooked during model training, SEG aims to make digital fabrication more sustainable and accessible. 
The primary reason SEG achieves consistent improvements is that ODPO reshapes the training signal distribution, amplifying meaningful preference differences in NSV. This prevents collapse in regions where support rewards are sparse. Future work will refine SEG and explore additional constraints for improved printability.

\section*{Acknowledgment}
The authors would thank the valuable help offered by the following students working in the Computational Robotics and Manufacturing Lab at the Chinese University of Hong Kong lead by Prof. Guoxin Fang, including Zhuo Huang and Richard Yixuan Zhou.

\bibliographystyle{IEEEtran}
\bibliography{reference}

@STRING{arxiv   = {arXiv preprint} }

@STRING{cvpr    = {Proc. IEEE Conf. Comput. Vis. Pattern Recognit. (CVPR)} }

@STRING{eccv    = {Proc. Eur. Conf. Comput. Vis. (ECCV)} }

@STRING{iccv    = {Proc.~of the IEEE/CVF Intl.~Conf.~on Computer Vision (ICCV)} }

@STRING{iclr    = {Proc.~of the Int.~Conf.~on Learning Representations (ICLR)}}

@STRING{icra    = {Proc. IEEE Int. Conf. Robot. Automat. (ICRA)} }

@STRING{neurips = {Proc.~of the Conference on Neural Information Processing Systems (NeurIPS)} }

@STRING{siggraph    = {Proc.~of the Intl.~Conf.~on Computer Graphics and Interactive Techniques (SIGGRAPH)} }

@inproceedings{dumas2014efficient,
  title={Efficient computation of support structures in 3D printing},
  author={Dumas, Jean and Lefebvre, Sylvain},
  booktitle={ACM SIGGRAPH 2014 Posters},
  pages={1--1},
  year={2014}
}

@article{vanek2014clever,
  title={Clever support: Efficient support structure generation for digital fabrication},
  author={Vanek, Jiri and Galicia, Jorge and Benes, Bedrich},
  journal={Computer Graphics Forum},
  volume={33},
  number={5},
  pages={117--125},
  year={2014},
  publisher={Wiley Online Library}
}

@inproceedings{wang2018supportnet,
  title={SupportNet: Efficient and accurate support generation for additive manufacturing},
  author={Wang, Tianhao and Gao, Shengnan and Wang, Lijia and Liu, Yasutaka and Liu, Ligang and Ju, Tao},
  booktitle={Computer Graphics Forum},
  volume={37},
  number={7},
  pages={177--187},
  year={2018},
  organization={Wiley Online Library}
}

@article{cap3d,
  title={Scalable 3d captioning with pretrained models},
  author={Luo, Tiange and Rockwell, Chris and Lee, Honglak and Johnson, Justin},
  journal={NeurIPS},
  volume={36},
  pages={75307--75337},
  year={2023}
}

@article{zhou2016thingi10k,
  title={Thingi10k: A dataset of 10,000 3d-printing models},
  author={Zhou, Qingnan and Jacobson, Alec},
  journal={arXiv preprint arXiv:1605.04797},
  year={2016}
}

@article{poole2022dreamfusion,
  title={DreamFusion: Text-to-3D using 2D diffusion models},
  author={Poole, Ben and Jain, Ajay and Barron, Jonathan T and Brooks, Dan B and Ziebart, Brian D and Mildenhall, Ben},
  journal={arXiv preprint arXiv:2209.14988},
  year={2022}
}

@article{GAO201565,
title = {The status, challenges, and future of additive manufacturing in engineering},
journal = {Computer-Aided Design},
volume = {69},
pages = {65-89},
year = {2015},
author = {Wei Gao and Yunbo Zhang and Devarajan Ramanujan and Karthik Ramani and Yong Chen and Christopher B. Williams and Charlie C.L. Wang and Yung C. Shin and Song Zhang and Pablo D. Zavattieri},
}

@article{VAISSIER201911,
title = {Genetic-algorithm based framework for lattice support structure optimization in additive manufacturing},
journal = {Computer-Aided Design},
volume = {110},
pages = {11-23},
year = {2019},
author = {Benjamin Vaissier and Jean-Philippe Pernot and Laurent Chougrani and Philippe Véron},
}

@article{Vanek14,
author = {Vanek, J. and Galicia, J. A. G. and Benes, B.},
title = {Clever Support: Efficient Support Structure Generation for Digital Fabrication},
journal = {Computer Graphics Forum},
volume = {33},
number = {5},
pages = {117-125},
year = {2014}
}

@article{Luo12,
author = {Luo, Linjie and Baran, Ilya and Rusinkiewicz, Szymon and Matusik, Wojciech},
title = {Chopper: partitioning models into 3D-printable parts},
year = {2012},
publisher = {Association for Computing Machinery},
address = {New York, NY, USA},
volume = {31},
number = {6},
journal = {ACM Trans. Graph.},
month = nov,
articleno = {129},
numpages = {9},
}

@article{Gaynor16,
	journal = {Structural and Multidisciplinary Optimization},
	number = {5},
    author = {Gaynor, Andrew T. and Guest, James K.},
	pages = {1157--1172},
	title = {Topology optimization considering overhang constraints: Eliminating sacrificial support material in additive manufacturing through design},
	volume = {54},
	year = {2016},
}

@article{amini2024direct,
  title={Direct preference optimization with an offset},
  author={Amini, Afra and Vieira, Tim and Cotterell, Ryan},
  journal={arXiv preprint arXiv:2402.10571},
  year={2024}
}

@article{hu2015support,
  title={Support slimming for single material based additive manufacturing},
  author={Hu, Kailun and Jin, Shuo and Wang, Charlie CL},
  journal={Computer-Aided Design},
  volume={65},
  pages={1--10},
  year={2015},
  publisher={Elsevier}
}

@article{Zhang2015,
 author = {Zhang, Xiaoting and Le, Xinyi and Panotopoulou, Athina and Whiting, Emily and Wang, Charlie C. L.},
 title = {Perceptual Models of Preference in 3D Printing Direction},
 journal = {ACM Trans. Graph.},
 issue_date = {November 2015},
 volume = {34},
 number = {6},
 month = oct,
 year = {2015},
 issn = {0730-0301},
 pages = {215:1--215:12},
 articleno = {215},
 numpages = {12},
 doi = {10.1145/2816795.2818121},
 acmid = {2818121},
 publisher = {ACM},
}

@article{Gao2015,
author = "Wei Gao and Yunbo Zhang and Devarajan Ramanujan and Karthik Ramani and Yong Chen and Christopher B. Williams and Charlie C.L. Wang and Yung C. Shin and Song Zhang and Pablo D. Zavattieri",
title = "The status, challenges, and future of additive manufacturing in engineering ",
journal = "Computer-Aided Design",
volume = "69",
pages = "65--89",
year = "2015",
}

@article{kingma2014adam,
  title={Adam: A method for stochastic optimization},
  author={Kingma, Diederik P and Ba, Jimmy},
  journal={arXiv preprint arXiv:1412.6980},
  year={2014}
}

@inproceedings{rasley2020deepspeed,
  title={Deepspeed: System optimizations enable training deep learning models with over 100 billion parameters},
  author={Rasley, Jeff and Rajbhandari, Samyam and Ruwase, Olatunji and He, Yuxiong},
  booktitle={Proceedings of the 26th ACM SIGKDD international conference on knowledge discovery \& data mining},
  pages={3505--3506},
  year={2020}
}

@inproceedings{wu2017robofdm,
  title={RoboFDM: A robotic system for support-free fabrication using FDM},
  author={Wu, Chenming and Dai, Chengkai and Fang, Guoxin and Liu, Yong-Jin and Wang, Charlie CL},
  booktitle={ICRA},
  pages={1175--1180},
  year={2017},
  organization={IEEE}
}

@inproceedings{wallace2024diffusion,
  title={Diffusion model alignment using direct preference optimization},
  author={Wallace, Bram and Dang, Meihua and Rafailov, Rafael and Zhou, Linqi and Lou, Aaron and Purushwalkam, Senthil and Ermon, Stefano and Xiong, Caiming and Joty, Shafiq and Naik, Nikhil},
  booktitle={CVPR},
  pages={8228--8238},
  year={2024}
}

@article{wald2014embree,
  title={Embree: a kernel framework for efficient CPU ray tracing},
  author={Wald, Ingo and Woop, Sven and Benthin, Carsten and Johnson, Gregory S and Ernst, Manfred},
  journal={ACM Trans. Graph.},
  volume={33},
  number={4},
  pages={1--8},
  year={2014},
  publisher={ACM New York, NY, USA}
}

@article{hu2020fast,
  title={Fast tetrahedral meshing in the wild},
  author={Hu, Yixin and Schneider, Teseo and Wang, Bolun and Zorin, Denis and Panozzo, Daniele},
  journal={ACM Trans. Graph.},
  volume={39},
  number={4},
  pages={117--1},
  year={2020},
  publisher={ACM New York, NY, USA}
}

@InProceedings{Lin_2023_CVPR,
    author    = {Lin, Chen-Hsuan and Gao, Jun and Tang, Luming and Takikawa, Towaki and Zeng, Xiaohui and Huang, Xun and Kreis, Karsten and Fidler, Sanja and Liu, Ming-Yu and Lin, Tsung-Yi},
    title     = {Magic3D: High-Resolution Text-to-3D Content Creation},
    booktitle = {CVPR},
    month     = {June},
    year      = {2023},
    pages     = {300-309}
}

@INPROCEEDINGS{Lorraine23,
  author={Lorraine, Jonathan and Xie, Kevin and Zeng, Xiaohui and Lin, Chen-Hsuan and Takikawa, Towaki and Sharp, Nicholas and Lin, Tsung-Yi and Liu, Ming-Yu and Fidler, Sanja and Lucas, James},
  booktitle={2023 IEEE/CVF International Conference on Computer Vision (ICCV)}, 
  title={ATT3D: Amortized Text-to-3D Object Synthesis}, 
  year={2023},
  volume={},
  number={},
  pages={17900-17910},
}

@article{xie2024latte3d,
  title = {LATTE3D: Large-scale Amortized Text-To-Enhanced3D Synthesis},
  author = {Kevin Xie and Jonathan Lorraine and Tianshi Cao and Jun Gao and James Lucas and Antonio Torralba and Sanja Fidler and Xiaohui Zeng},
  journal = {ECCV},
  year = {2024},
}

@unknown{Jun23,
author = {Jun, Heewoo and Nichol, Alex},
year = {2023},
month = {05},
pages = {},
title = {Shap-E: Generating Conditional 3D Implicit Functions},
doi = {10.48550/arXiv.2305.02463}
}

@article{tang2023dreamgaussian,
  title={DreamGaussian: Generative Gaussian Splatting for Efficient 3D Content Creation},
  author={Tang, Jiaxiang and Ren, Jiawei and Zhou, Hang and Liu, Ziwei and Zeng, Gang},
  journal={arXiv preprint arXiv:2309.16653},
  year={2023}
}

@inproceedings{wang2023prolificdreamer,
  title={ProlificDreamer: High-Fidelity and Diverse Text-to-3D Generation with Variational Score Distillation},
  author={Zhengyi Wang and Cheng Lu and Yikai Wang and Fan Bao and Chongxuan Li and Hang Su and Jun Zhu},
  booktitle={NeurIPS},
  year={2023}
}

@INPROCEEDINGS{Tsalicoglou24,
  author={Tsalicoglou, Christina and Manhardt, Fabian and Tonioni, Alessio and Niemeyer, Michael and Tombari, Federico},
  booktitle={2024 International Conference on 3D Vision (3DV)}, 
  title={TextMesh: Generation of Realistic 3D Meshes From Text Prompts}, 
  year={2024},
  volume={},
  number={},
  pages={1554-1563},
}

@inproceedings{gao2022get3d,
title={GET3D: A Generative Model of High Quality 3D Textured Shapes Learned from Images},
author={Jun Gao and Tianchang Shen and Zian Wang and Wenzheng Chen and Kangxue Yin and Daiqing Li and Or Litany and Zan Gojcic and Sanja Fidler},
booktitle={NeurIPS},
year={2022}
}

@article{nichol2022point,
  title={Point-e: A system for generating 3d point clouds from complex prompts},
  author={Nichol, Alex and Jun, Heewoo and Dhariwal, Prafulla and Mishkin, Pamela and Chen, Mark},
  journal={arXiv preprint arXiv:2212.08751},
  year={2022}
}

@article{xiang2024structured,
  title={Structured 3D Latents for Scalable and Versatile 3D Generation},
  author={Xiang, Jianfeng and Lv, Zelong and Xu, Sicheng and Deng, Yu and Wang, Ruicheng and Zhang, Bowen and Chen, Dong and Tong, Xin and Yang, Jiaolong},
  journal={arXiv preprint arXiv:2412.01506},
  year={2024}
}

@article{chen2024atlas3d,
      title={Atlas3D: Physically Constrained Self-Supporting Text-to-3D for Simulation and Fabrication},
      author={Yunuo Chen and Tianyi Xie and Zeshun Zong and Xuan Li and Feng Gao and Yin Yang and Ying Nian Wu and Chenfanfu Jiang},
      journal={arXiv preprint arXiv:2405.18515},
      year={2024},
}

@article{li2025dso,
    title   = {DSO: Aligning 3D Generators with Simulation Feedback for Physical Soundness},
    author  = {Li, Ruining and Zheng, Chuanxia and Rupprecht, Christian and Vedaldi, Andrea},
    journal = {arXiv preprint arXiv:2503.22677},
    year    = {2025}
}

@article{wu2019general,
  title={General support-effective decomposition for multi-directional 3-D printing},
  author={Wu, Chenming and Dai, Chengkai and Fang, Guoxin and Liu, Yong-Jin and Wang, Charlie CL},
  journal={IEEE Transactions on Automation Science and Engineering},
  volume={17},
  number={2},
  pages={599--610},
  year={2019},
}

@book{norrie2014finite,
  title={The finite element method: fundamentals and applications},
  author={Norrie, Douglas H and De Vries, Gerard},
  year={2014},
  publisher={Academic Press}
}

@article{hu2022lora,
  title={Lora: Low-rank adaptation of large language models.},
  author={Hu, Edward J and Shen, Yelong and Wallis, Phillip and Allen-Zhu, Zeyuan and Li, Yuanzhi and Wang, Shean and Wang, Lu and Chen, Weizhu and others},
  journal={ICLR},
  volume={1},
  number={2},
  pages={3},
  year={2022}
}

@article{dai2018support,
  title={Support-free volume printing by multi-axis motion},
  author={Dai, Chengkai and Wang, Charlie CL and Wu, Chenming and Lefebvre, Sylvain and Fang, Guoxin and Liu, Yong-Jin},
  journal={ACM Trans. Graph.},
  volume={37},
  number={4},
  pages={1--14},
  year={2018},
  publisher={ACM New York, NY, USA}
}

@article{jang2020free,
  title={Free-floating support structure generation},
  author={Jang, Seongje and Moon, Byungjin and Lee, Kunwoo},
  journal={Computer-Aided Design},
  volume={128},
  pages={102908},
  year={2020},
  publisher={Elsevier}
}

@inproceedings{radford2021learning,
  title={Learning transferable visual models from natural language supervision},
  author={Radford, Alec and Kim, Jong Wook and Hallacy, Chris and Ramesh, Aditya and Goh, Gabriel and Agarwal, Sandhini and Sastry, Girish and Askell, Amanda and Mishkin, Pamela and Clark, Jack and others},
  booktitle={International conference on machine learning},
  pages={8748--8763},
  year={2021},
  organization={PmLR}
}



\end{document}